# Classification of Disease from Lungs X-ray Images using VGG16, VGG19 and ResNet50 Models

**[1]Nand Lal Yadav, [1]Rajesh Kumar, [1]Satyendra Singh, [1*]Sudhakar Singh**

[1]Department of Electronics and Communication, University of Allahabad, Prayagraj, India.

[*]*Corresponding Author*

*E-mail addresses:*

nly.jct@gmail.com(Nand Lal Yadav)

rajeshkumariitbhu@gmail.com(Rajesh Kumar)

satyendra@allduniv.ac.in(Satyendra Singh)

sudhakar@allduniv.ac.in (Sudhakar Singh)

***Abstract***

With the increase in the number of cases related to respiratory diseases, there is an urgent need to detect them early and diagnose them accurately. Convolutional neural networks have given promising results when used for diagnosing diseases using imaging tests. In this study, we investigate the potential of applying deep learning algorithms such as VGG16, VGG19, and ResNet50 for classification of lung ailments based on X-ray images. A detailed analysis of the aforementioned models' performances was conducted to assess how well they can classify various types of lung ailments, including pneumonia, tuberculosis, lung cancer, and normal lungs. In order to do that, these deep learning models were trained on a vast amount of X-ray images. The results of our study show that while all three models provide good results, ResNet-50 performs best in comparison with other models due to its efficiency and high level of accuracy. We believe that these deep learning models can be successfully implemented in the practice of diagnosing pulmonary diseases in the future. It helps with early disease detection and improves patient outcomes.

## 1. INTRODUCTION

There is an enormous number of deaths and sickness related to the lung disease, and it is a global health problem. Lung Disease Disorder refers to the wide range of disorders related to but not limited to pneumonia, tuberculosis (TB), chronic obstructive lung disease (COPD), and lung cancer. Lung diseases include different kinds of lung diseases, as shown in Table 1. There are only a few cases of lung disease, and there are countless other possibilities where the breathing process can be affected.

Table 1: Types of lung diseases and their characteristics

| Sl. No. | Lung Disease | Characteristics |
|---|---|---|
| 1. | Pneumonia | Infecting this disease leads to inflammation in the air sacs of the lungs that have the symptoms of fever, cough, and difficulty in breathing. |
| 2. | Tuberculosis (TB) | The TB disease is a bacterial infection which involves the lungs and is caused by Mycobacterium tuberculosis. Its symptoms include chronic coughing, chest pains, and weight loss. |
| 3. | Lung Cancer | This is a type of cancer where abnormal cells form in the lung and its most common cause is smoking. Its symptoms include coughing, blood-streaked phlegm, and chest pain. |
| 4. | Chronic Obstructive Pulmonary Disease (COPD) | Emphysema and chronic bronchitis are some examples of conditions that are included in the Chronic Obstructive Pulmonary Disease. This causes airway irritation that causes reduced breath and chronic coughing. |
| 5. | Asthma | Asthma is a chronic respiratory illness characterized by airway irritation and increased reactivity, resulting in recurrent episodes of wheezing, coughing, and breathlessness. |
| 6. | Interstitial Lung Disease (ILD) | ILD encompasses a group of disorders which create scarring of the lung tissue, affecting its ability to expand and contract properly. |
| 7. | Pulmonary Hypertension | This condition involves increased blood pressure in the pulmonary arteries, showing symptoms like shortness of breath, tiredness, and chest pain. |
| 8. | Cystic Fibrosis | Cystic fibrosis is a genetic disorder harming the lungs and other organs, leading to the growth of thick and sticky mucus, causing respiratory issues. |
| 9. | Lung Fibrosis | Lung fibrosis refers to the scarring of lung tissue due to various causes, resulting in reduced lung function and breathlessness. |
| 10. | Lung Abscess | Lung abscess is defined as a localized collection of pus that develops within the tissue of the lungs. It can be caused by various factors, such as a bacterial infection. |

Early diagnosis and effective treatment of respiratory diseases can be achieved through screening. It has been realized that the suitability of certain screening practices hinges on such factors as medical history, risk factors, and symptoms of an individual. These factors should be taken into consideration by healthcare practitioners in their recommendations of screening tests. Early diagnosis with screening has the potential of greatly enhancing the treatment outcomes and value of life in respiratory disease patients.

Respiratory disease screening is important to detect and provide effective treatment. There are a number of screening systems for different lung conditions [5]. The chest X-rays are usually taken to identify the abnormalities of the lung tissue and are useful in detecting the diseases such as pneumonia, tuberculosis, lung cancer and lung abscesses. Computed Tomography (CT) Scans provide a clearer picture and are necessary to detect lung cancer, lung fibrosis and interstitial lung disease. Tuberculosis and infections can be diagnosed using a sputum test through the examination of mucus samples. Pulmonary functional tests like spirometry can measure lung functions, which is used in the diagnosis of asthma and COPD [6]. Bronchoscopy enables the airways to be seen and lung cancer to be diagnosed, as well as taking samples. Definitive diagnosis sometimes requires lung biopsies. Screening programs, including low-dose CT scans, are implemented in high-risk groups in some cases to detect lung cancer in people who have a history of smoking.

Screenings are determined by an individual's symptoms, medical history and risk factors. Screenings can help in improving the outcome and quality of life of patients with lung diseases, as they can be treated early. Regular screening, especially in those with risk factors, is essential for the prevention of lung diseases. Computer-aided diagnosis (CAD) and the methodology of machine learning are powerful tools in medical imaging, and particularly in the diagnosis of respiratory diseases by X-ray images [7]. They are helpful as they can recognise patterns, be efficient in analysing images and provide quantitative measurements. Machine learning models are trained on huge datasets, enabling them to learn to detect subtle features not usually recognised by the human eye. This results in faster and more accurate image analysis. Furthermore, the degree of quantitative assessment offered by the models can help in the precise measurement of lesion characteristics to help distinguish benign from malignant tumours (including pulmonary nodules) [8]. Machine learning algorithms and computer-assisted detection can help avoid human errors, offer an unbiased and uniform method of image processing, and assist in making early diagnoses.

These diseases may be difficult to identify and differentiate based on X-ray images as they have slight differences in patterns, and the radiological features may overlap. Efficient patient care and treatment planning are dependent on timely and correct diagnosis of such diseases. Medical images, including chest X-rays, interpretation has been a fundamental component in disease diagnosis since time immemorial [1]. Nonetheless, this process may be laborious and prone to human errors. A specialized branch of artificial intelligence, deep learning has received considerable recognition in the healthcare field due to its ability to transform the diagnosis and treatment of various diseases, including those of the lungs [2]. The deep learning models and especially Convolution Neural Networks (CNNs) have shown impressive performance in image analysis activities, and hence, it is an effective method in automatic classification of lung diseases through X-ray images. CNNs interpret the features of images, which are important and give predictions that are based on the learned information [3]. The CNN models that have been shown to perform well in a number of image recognition challenges are VGG16, VGG19 and ResNet-50 [4]. These models have a major benefit of transfer learning. They have been trained for general image classification jobs on large data sets, and can be fine-tuned on medical image investigation tasks with a relatively smaller volume of data from diseases investigated. This is particularly advantageous when it may be difficult to obtain large amounts of medical data. High accuracy is also a feature of VGG16, VGG19 and ResNet-50. They have a high accuracy in image classification, which helps to improve the diagnostic accuracy, which in turn increases the trust and accuracy of health care.

This paper aims to look deeper into the use of these advanced CNN models in the classification of lung diseases through X-rays. This paper compares the VGG16, VGG19 and ResNet-50 models in this important medical application. In particular, this paper compares the performance of VGG16, VGG19 and ResNet-50 CNN models in the classification of lung diseases like pneumonia, TB, lung cancer and normal lung on the large dataset of chest X-ray images.

This is an important work since it may help health care workers, particularly radiologists, to diagnose lung diseases. It is a step toward the usage of artificial intelligence in the medical image analysis and management of lung diseases. The process of lung disease detection and classification can be automated, which will aid in enhancing the diagnostic accuracy, efficiency and patient decision. Although such approaches are very promising, they may be most helpful and efficient when used along with the expertise of health workers, who can interpret the findings with references to the history of the patient. It is a team approach that involves human and technological skills that will result in perfection of diagnostic accuracy and will eventually benefit the patient in the process of lung disease detection.

## 2. LITERATURE REVIEW

The lung diseases that are a massive health burden to the world are pneumonia, TB and lung cancer. Interpretation of X-ray images manually is one of the conventional methods of diagnosis, which is time consuming and prone to human error. To counter this, the researchers have resorted to deep learning in ensuring that they create automated classification systems that can be used to identify and detect different pathologies of the lungs with accuracy [11].

Deep learning methods have been a new technology in the analysis of medical images, that is, the diagnosis of lung diseases using X-ray images. The high rates of pulmonary diseases show that proper and timely diagnosis is critical, and that is why the researchers are interested in using such advanced technologies as Convolutional Neural Networks (CNNs) that can automate the diagnostic process and make it efficient.

A number of studies have applied deep learning models to medical imaging, where they have realized its capability to transform the way diagnosis is done [9]. CNNs have demonstrated exceptional features in extracting features and recognizing patterns, and thus, they are highly applicable in activities that require intricate visual information, including medical images [10].

VGG architectures, as introduced by Simonyan and Zisserman [12], provided a base on which further developments in the sphere of deep learning would be achieved. With Deep convolutional layers, VGG16 and VGG19 were first used in image classification tasks and proved to be effective. Simultaneously, Residual Networks (ResNets), namely ResNet-50, was able to solve the issue of vanishing gradients, resulting in better model performance and training efficiency [13].

The relevance of strong datasets can not be overrated in designing and assessing deep learning-based medical imaging. Irvin et al. [14] have used large volumes of X-ray images that have mixed cases and include pneumonia, tuberculosis, lung cancer, and normal lung conditions.

These datasets will be used to train and test the models, making sure that they can generalize to a range of pathologies.

Converting the deep learning models in medical practice out of research is a significant milestone in the achievement of the impact of deep learning models on patient outcomes. Esteva et al. [16] have discussed the possible clinical implications of these models, and they are used as aids to radiologists and other health care providers. These models have the potential to transform patient care, as they can make the diagnostic process quicker and more accurate.

ResNet, a concept suggested by He et al. in 2016 [17], has brought about the concept of residual learning, which enables extremely deep neural networks to be trained. One type of the ResNet architecture, ResNet50, has become popular due to its capability to crack the vanishing gradient problem and to successfully train very deep networks.

The VGG16, VGG19, and ResNet-50, as standard CNN structures, are essential in the recognition of respiratory diseases, based on X-ray, and transformed the field of medical image research. These models are good at extracting features, automatically discovering intricate patterns in lung X-ray images, which is crucial in detecting disease-related features and abnormalities. These pre-trained networks can be fine-tuned on labeled X-ray datasets to effectively classify diseases like pneumonia, tuberculosis, lung cancer and others.

Research like Alshmrani et al. [18] has made use of VGG19 in classifying lung diseases in X-ray images, demonstrating superiority in dealing with complex hierarchies of features. The authors suggested a deep learning model to classify Lung cancer, TB, Lung Opacity, and COVID-19. They employed VGG19+CNN i.e. VGG19 + three CNN blocks. In their research study, Khan and Aslam [15] have adopted VGG16, VGG19, ResNet50 and DenseNet121 deep learning models and transfer learning to detect Covid19 using X-ray images. A binary classification and multi-class classification model with DarkNet model 17 convolution layers with various filtering on each layer, was proposed by Ozturk et al. [28]. The binary classification accuracy was 98.08%, and the accuracy of multi-class classification was 87.02%.

The results and procedures of the state-of-the-art studies form an effective foundation of the current study, which will further streamline and broaden the applicability of deep learning in the lungs' X-ray images. VGG16, VGG19 and ResNet-50 are central elements of automated respiratory disease detection of X-ray images. Their feature extraction ability, classification ability, efficiency, adaptability, and interpretability have transformed the diagnostic process and resulted in improved patient outcomes and simplified the healthcare procedure. These algorithms represent an important breakthrough in the utilization of artificial intelligence potential in the medical imaging and respiratory disease diagnosis domain.

## 3. MATERIALS AND METHODS

The covid19-image-dataset[1] is a collection of medical images related to respiratory conditions, particularly COVID-19. The dataset is divided into two subsets, i.e. "train" and "test". It includes images categorized into "Covid," "Normal, “and "Viral Pneumonia”. A general framework of the classification is depicted in Fig. 1. The training set contains images for each category, allowing machine learning models to learn patterns connected with different respiratory conditions. The dataset is likely intended for training and evaluating models for the auto-classification of X-ray images, aiding in the diagnosis of respiratory conditions, including COVID-19. Ethical considerations regarding medical data privacy should be considered when using such datasets.

---

[1] https://www.kaggle.com/datasets/prashant268/chest-xray-covid19-pneumonia/download?datasetVersionNumber=2

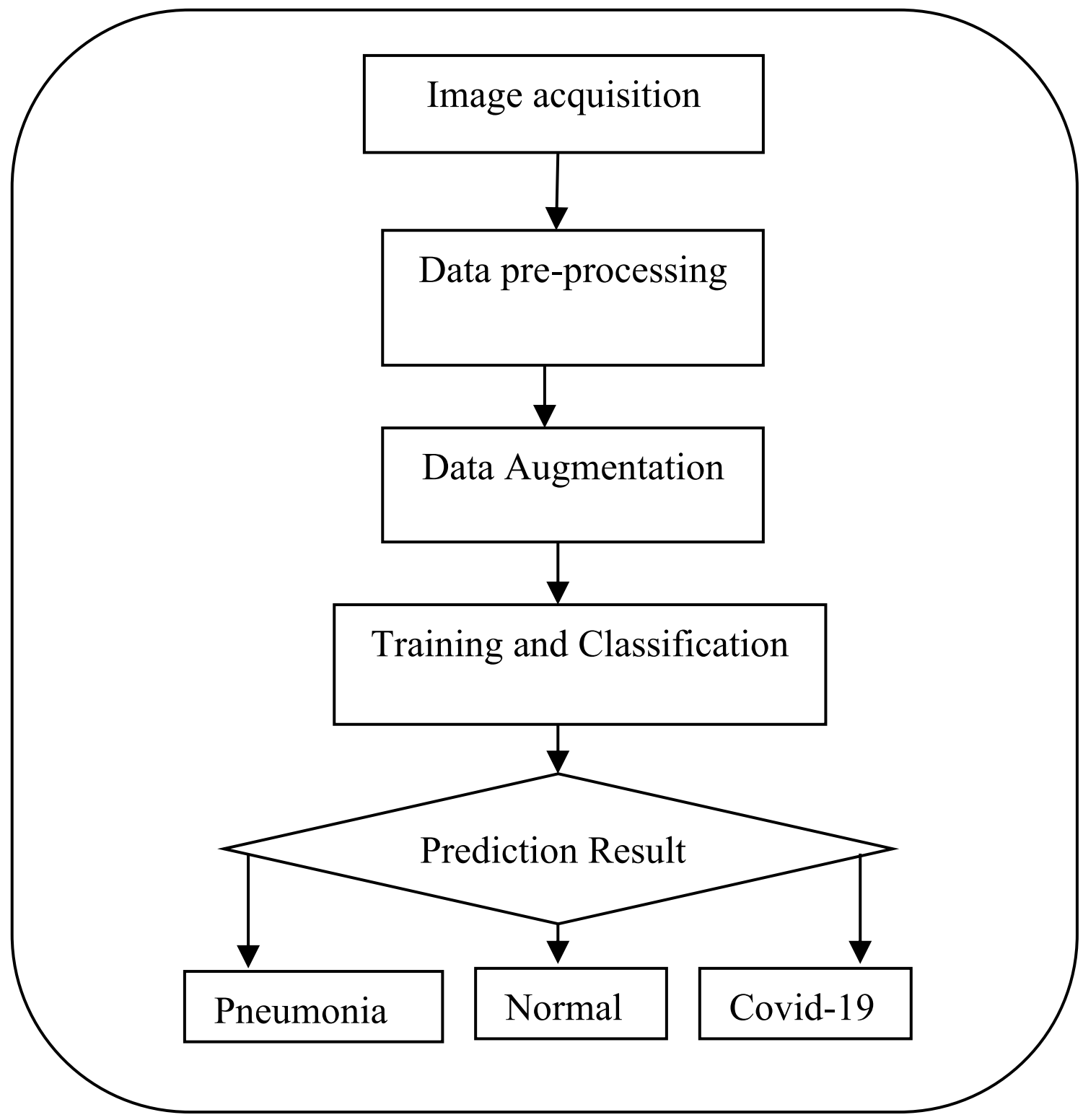

Fig.1: Framework of the proposed model

### 3.1 Augmentation Process

The provided parameters are part of an augmentation process commonly used in the training of deep learning models designed for computer vision tasks, especially with image data [19]. Augmentation is a technique where the input data is artificially extended by applying numerous transformations to the source images. This helps enhance the model's generality by exposing it to a broader range of variations and orientations of the input data. The details of each parameter used in our experiment have been described as follows.

*rescale=Image I/255:* This parameter scales the pixel values of the images. Rescaling is a standard preprocessing step, and dividing by 255 is common for images having pixel values in the range of 0 to 255. It helps bring the pixel values in a range from 0 to 1.

*samplewise_center=True:* Centres each sample (image) by subtracting the mean pixel value of that sample. This can help in reducing the impact of varying lighting conditions in the training dataset.

*samplewise_std_normalization=True:* Divides each input by its standard deviation value across all pixels. This step normalizes the data and helps in achieving a consistent scale for the features.

*rotation_range=40:* Randomly do rotation of the images by a degree in the range of -40 to +40 degrees. This helps the model to be more robust to different orientations of the objects in the images.

*width_shift_range=0.2:* Randomly shifts the images in a horizontal way by a fraction of the total width, in this case by up to 20%. This is useful for making the model invariant to small translations.

*height_shift_range=0.2:* Similar to width shift, this parameter randomly shifts the images in a vertical way by a fraction of the total height, up to 20%.

*shear_range=0.2:* Randomly applies shearing transformations to the images. Shearing distorts the shape of objects, and this parameter controls the intensity of the shearing.

*zoom_range=0.2:* Images are randomly zoomed by a factor of up to 20%. This helps the model to learn to recognize objects at different scales.

*horizontal_flip=True:* Randomly flips the images in a horizontal direction. This is a common augmentation technique as it helps the model to become invariant to the orientation of objects.

*cval=0:*This value is used for the points outside the boundaries when filling the image after transformations. It is set to 0, meaning black pixels will be used.

*validation_split=0.2:* This parameter is not an augmentation technique but rather a way to split the dataset into training and validation sets. It specifies the data fraction to be used for validation. In this case, 20% of the data will be utilized for validation purposes and the remaining 80% for training.

## 3.2 Convolutional Neural Network (CNN) Models

In recent years, CNN models, which are famous models of deep learning techniques, have shown demonstrated impressive performance in medical image analysis tasks, including segmentation for meaningful features. An accurate and efficient segmentation of the lung is essential for the diagnosis, treatment and monitoring of diseases in the lung. Deep learning methods are able to automatically extract pertinent features from medical photos and generate precise segmentation outcomes.

VGG16, VGG19, & ResNet-50 are the most popular CNN architectures that are used in deep learning enabled computer vision tasks, particularly in image feature extraction and classification. Here's a brief overview of each model.

### *3.2.1 VGG16*

VGG16 is a CNN architecture. The Visual Geometry Group (VGG) was developed by the University of Oxford. It is known for its simplicity and effectiveness. The "16" in its name refers to the 16 weight layers it comprises, which include 13 convolutional layers and 3 fully connected layers. VGG16 is specified by its use of small 3x3 convolutional filters and max-pooling layers, which contribute to its ability to learn rich image features. Despite its depth,

VGG16 is relatively easy to understand and has been extensively used as a building block in many deep learning applications [20].

*3.2.2VGG19*

The same research group develops VGG19. It has 19 layers of weight, a bit deeper. The extra layers of VGG19 are designed to extract more complex details and features of images. Similar to VGG16, VGG19 uses small convolutional filters, and max-pooling layers, which is why it performs well in diverse image recognition exercises [21].

*3.2.3ResNet-50v2*

Residual Network 50 version 2 (ResNet-50v2) is a deep convolutional neural network model, founded on the original ResNet-50, to overcome some limitations. ResNet-50v2 (introduced by Microsoft Research) adds important characteristics that enable it to be very useful in image recognition tasks [22]. The architecture is based on the concept of residual blocks, comprising skip connections or shortcuts. The shortcuts allow the network to learn the residual functions, which reduces the vanishing gradient problem and allows very deep networks to be trained. ResNet-50v2 consists of 50 layers, with each having a convolutional, batch normalization, and ReLU (Rectified Linear Unit) activation. An interesting aspect of ResNet-50v2 is that it has a bottleneck structure in the residual blocks. The design consists of 1x1 convolutional layers of feature reduction and restoration, and a 3x3 convolutional layer. This will improve the efficiency of computations. The convolutional layer is applied with the use of batch normalization, that also helps to stabilize and accelerate the training process. To achieve smaller representations, global average pooling (GAP) is used in place of fully connected layers, and the resultant reduction of the spatial dimensions causes the last classification layer. With shortcut connections, the gradients flow more efficiently during the backpropagation process, and this facilitates the process of training deep networks. ResNet-50v2 also has a weight initialization, He initialization, which improves the convergence during training. The implementation of the network can be found in popular deep learning models like TensorFlow and PyTorch. ResNet-50v2 in pre-trained form is commonly used by researchers and practitioners in a assortment of computer vision tasks, for example, image classification and segmentation, object detection [23]. The abilities of these models to extract and learn more complex image features have made them more applicable to the tasks of image classification and object detection, etc., and they are frequently used as a basis by researchers and practitioners and tailored to their particular deep learning tasks [24].

### 3.3 Performance Evaluation Metrics

The performance measures of the models used in the study are calculated using the following parameters [25][26][27][29].

*3.3.1 Confusion Matrix*

The confusion matrix provides an elaborate analysis of the predictions and true classes. The four elements of the confusion matrix are described as follows.

True Positive (TP): Cases that have been predicted positively. These are cases that have been predicted by the algorithm as positives.

True Negative (TN): Cases that have been predicted negatively. These are cases that have been predicted by the algorithm as negatives.

False Positive (FP): Instances which are incorrectly forecasted as positive. These are those cases where the model predicted the positive class, but the actual classes were negative. It is a Type I error.

False Negative (FN): Instances which are incorrectly forecasted as negative. These are those cases where the model forecast the negative class, but the actual classes were positive. It is a Type II error.

*3.3.2 Accuracy*

It is used to report the overall accuracy of each model in classifying lung diseases. This metric reflects the fraction of correctly classified cases out of the total.

*3.3.3 Sensitivity and Specificity*

Sensitivity (recall) indicates the capability of the model to correctly recognize the positive cases (e.g., diseased lungs), while specificity reflects its ability to classify the negative cases correctly (e.g., normal lungs).

*3.3.4 Precision*

Precision measures the accurateness of positive predictions. It indicates the fraction of correctly identified positive cases out of all predicted positives.

*3.3.5 F1 Score*

The F1 score is the harmonic mean of the value of precision and recall. It provides a stable measure of a model's performance.

*3.3.6 Training and Validation Loss*

The curves of the training and validation loss illustrate how well the models are learning over time. A decreasing loss indicates effective learning.

## 4 EXPERIMENTAL RESULTS

The experimentation is carried out with VGG16, VGG19, and ResNet-50v2 models with and without augmentation for the classification of diseases for X-ray images of the lung. Fig. 2, 3, and 4 show the performance variation without augmentation, while Fig. 5, 6, and 7 show the performance variation with augmentation. In all these figures, the number of epochs is represented by the x-axis, and accuracy and loss are represented by the y-axis.

Table 2 displays the outcomes derived from assessing the VGG16, VGG19, and ResNet-50v2models without augmentation. It encapsulates various performance metrics crucial for evaluating the efficiency of the models. Loss, Accuracy, Precision, Recall, and F1 score are among the performance evaluation criteria. Table 3 presents the results with Loss, Accuracy, Precision, Recall, and F1 scores of the multiclass classification with augmentation.

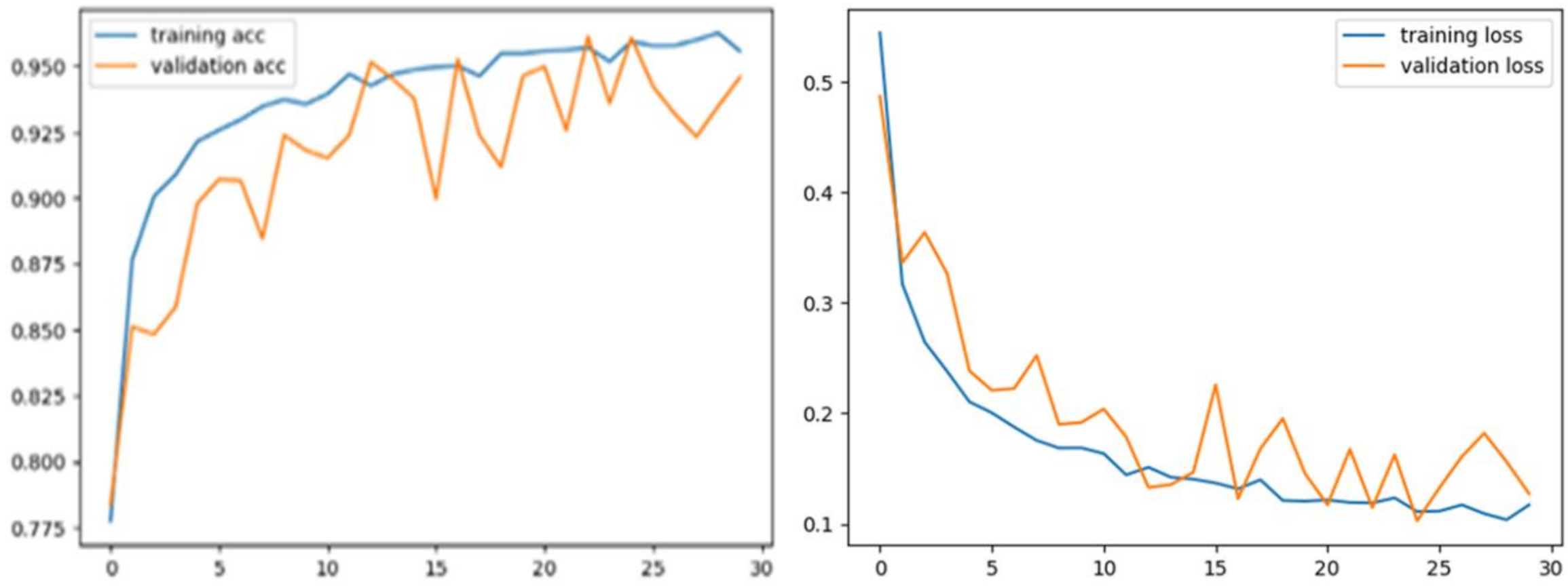

Fig. 2: Performance variations of the VGG16 model without augmentation

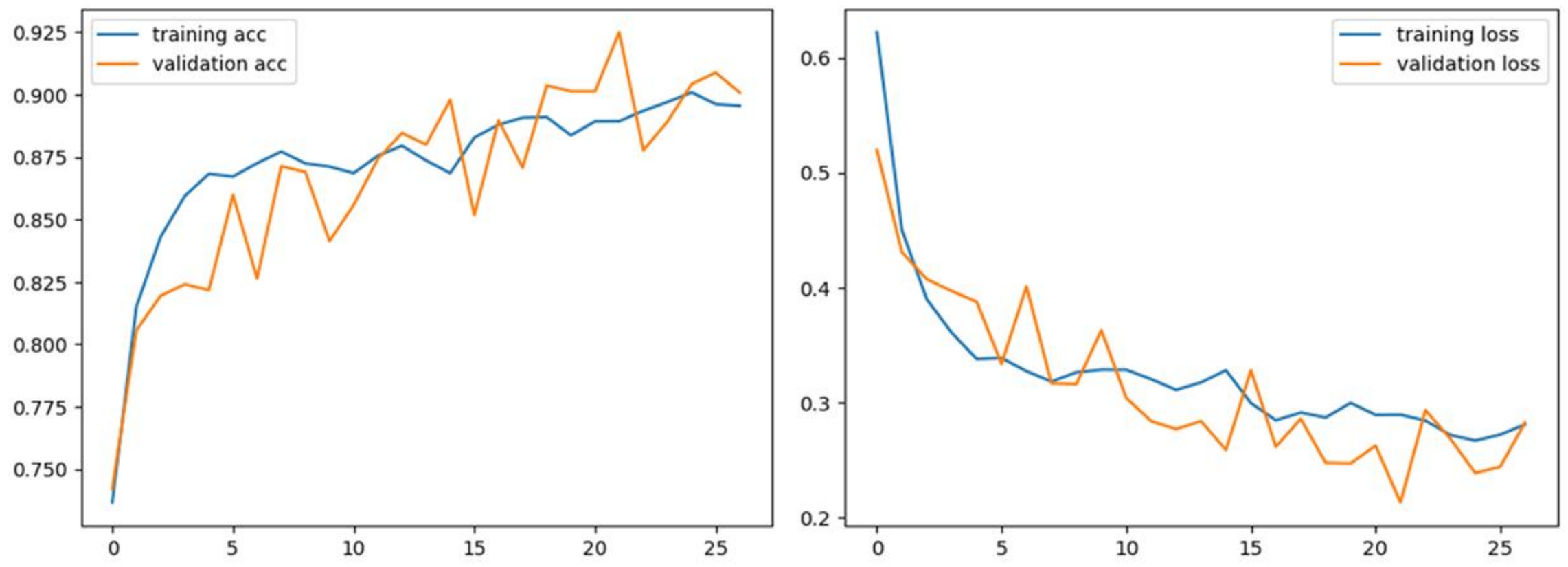

Fig. 3: Performance variation of the VGG19 model without augmentation

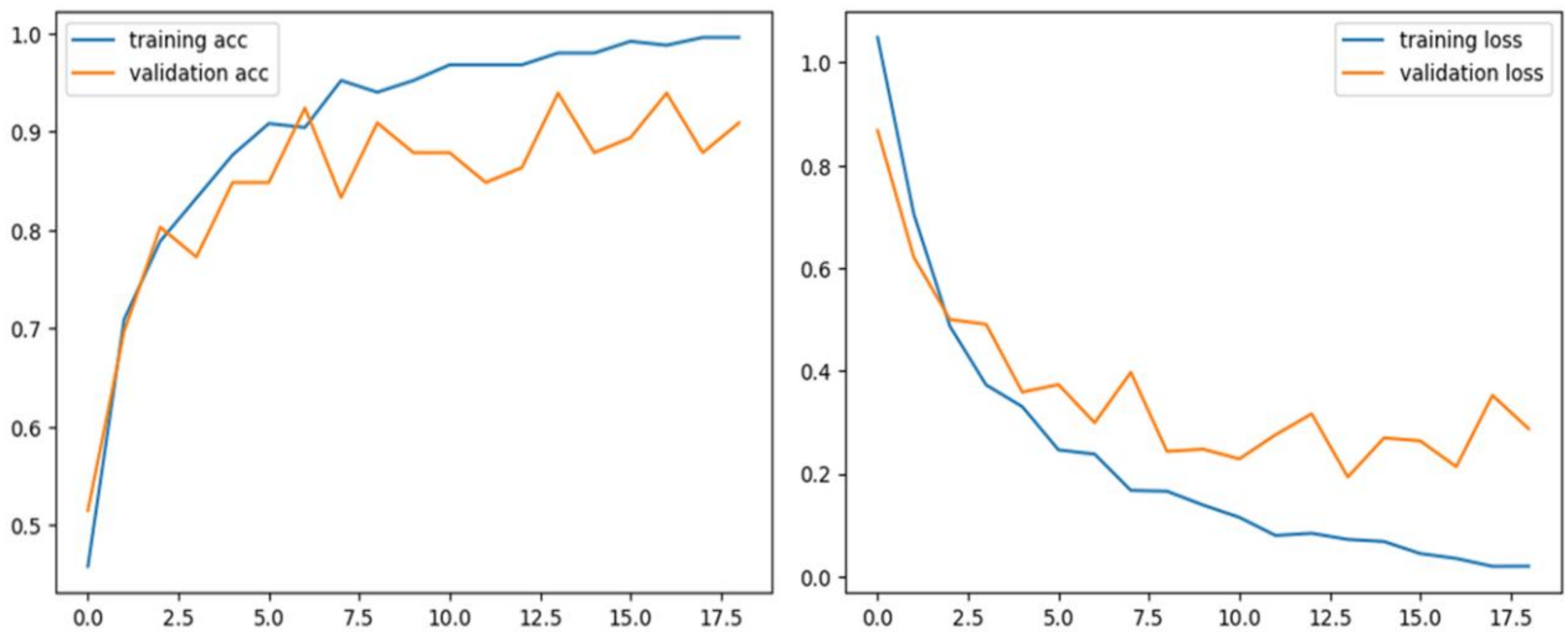

Fig. 4: Performance variation of the ResNet50v2 model without augmentation

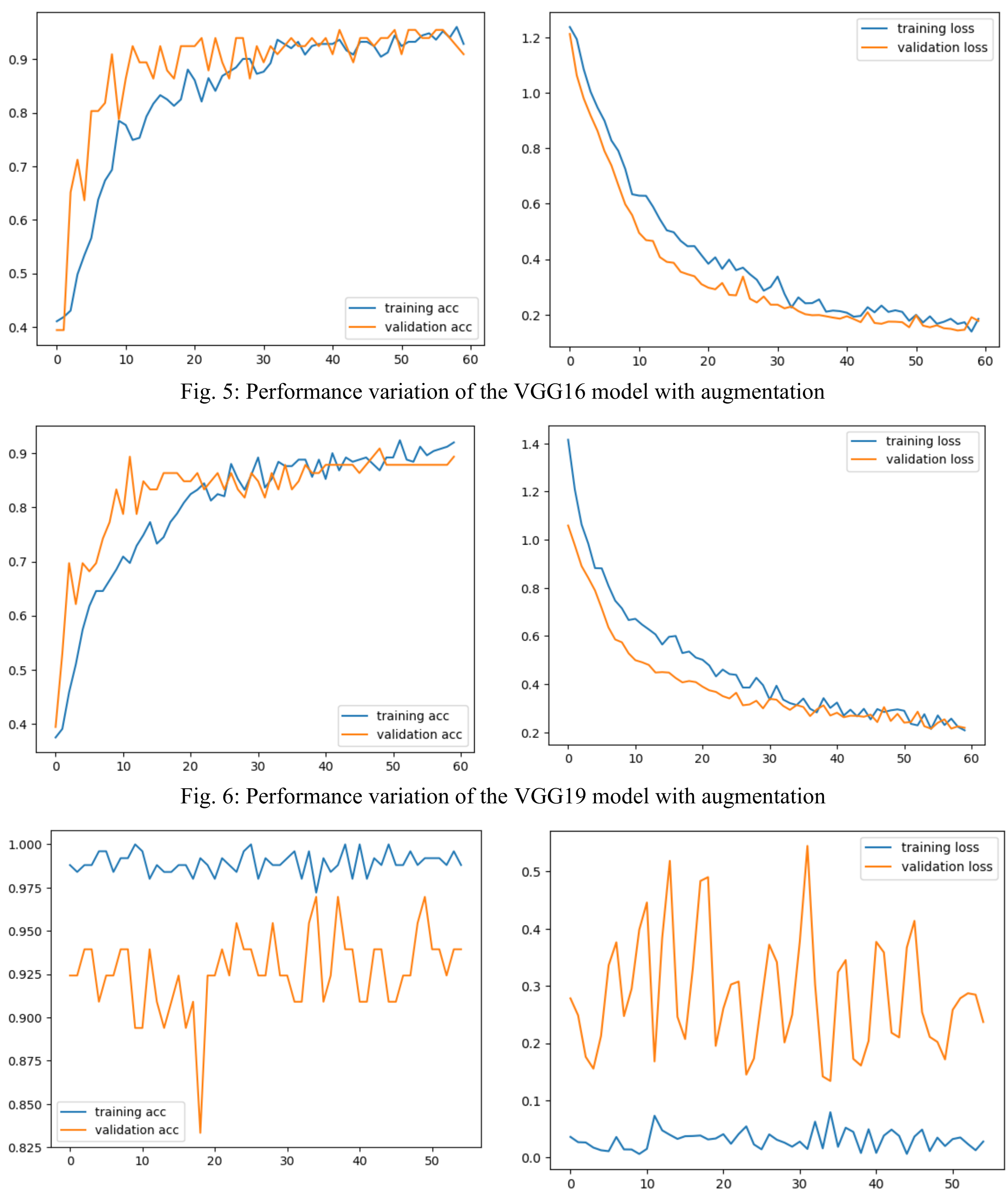

Fig. 5: Performance variation of the VGG16 model with augmentation

Fig. 6: Performance variation of the VGG19 model with augmentation

Fig. 7: Performance variation of the ResNet50v2model with augmentation

The results of the multiclass classifier models have been observed from the experimental analysis, and the multiclass classifier's confusion matrix is shown in figures Fig. 8, 9, and 10.

Table 2: Experimental results without augmentation

| Model | Loss | Accuracy | Class | Precision | Recall | F1 score |
|---|---|---|---|---|---|---|
| VGG16 | 0.1027 | 0.9607 | Normal | 0.96 | 0.98 | 0.97 |
| | | | Pneumonia | 0.99 | 0.94 | 0.96 |
| | | | Covid | 0.94 | 0.91 | 0.92 |
| VGG19 | 0.2128 | 0.9249 | Normal | 0.93 | 0.96 | 0.95 |
| | | | Pneumonia | 1.00 | 0.92 | 0.96 |
| | | | Covid | 0.88 | 0.81 | 0.85 |
| ResNet-50v2 | 0.19 | 0.93 | Normal | 0.95 | 0.90 | 0.92 |
| | | | Pneumonia | 0.87 | 1.00 | 0.93 |
| | | | Covid | 1.00 | 0.92 | 0.96 |

Table 3: Experimental results with augmentation

| Model | Loss | Accuracy | Class | Precision | Recall | F1 score |
|---|---|---|---|---|---|---|
| VGG16 | 0.14 | 0.94 | Normal | 0.93 | 0.99 | 0.96 |
| | | | Pneumonia | 0.98 | 0.88 | 0.92 |
| | | | Covid | 0.97 | 0.83 | 0.89 |
| VGG19 | 0.32 | 0.85 | Normal | 0.84 | 0.97 | 0.90 |
| | | | Pneumonia | 1.00 | 0.81 | 0.90 |
| | | | Covid | 0.85 | 0.51 | 0.64 |
| ResNet-50v2 | 0.13 | 0.95 | Normal | 0.95 | 0.95 | 0.95 |
| | | | Pneumonia | 0.91 | 1.00 | 0.95 |
| | | | Covid | 1.00 | 0.92 | 0.96 |

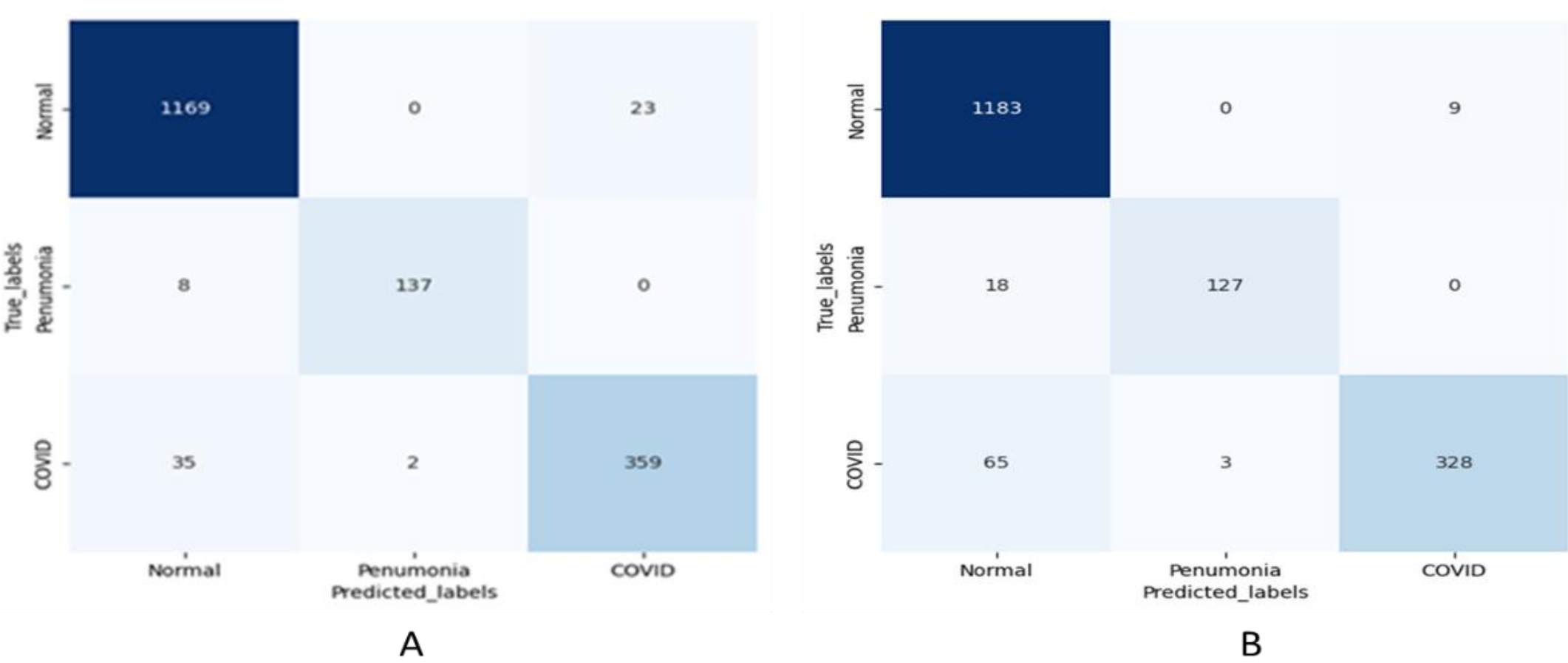

Fig. 8: VGG16 model confusion matrix (A) without augmentation and (B) with augmentation

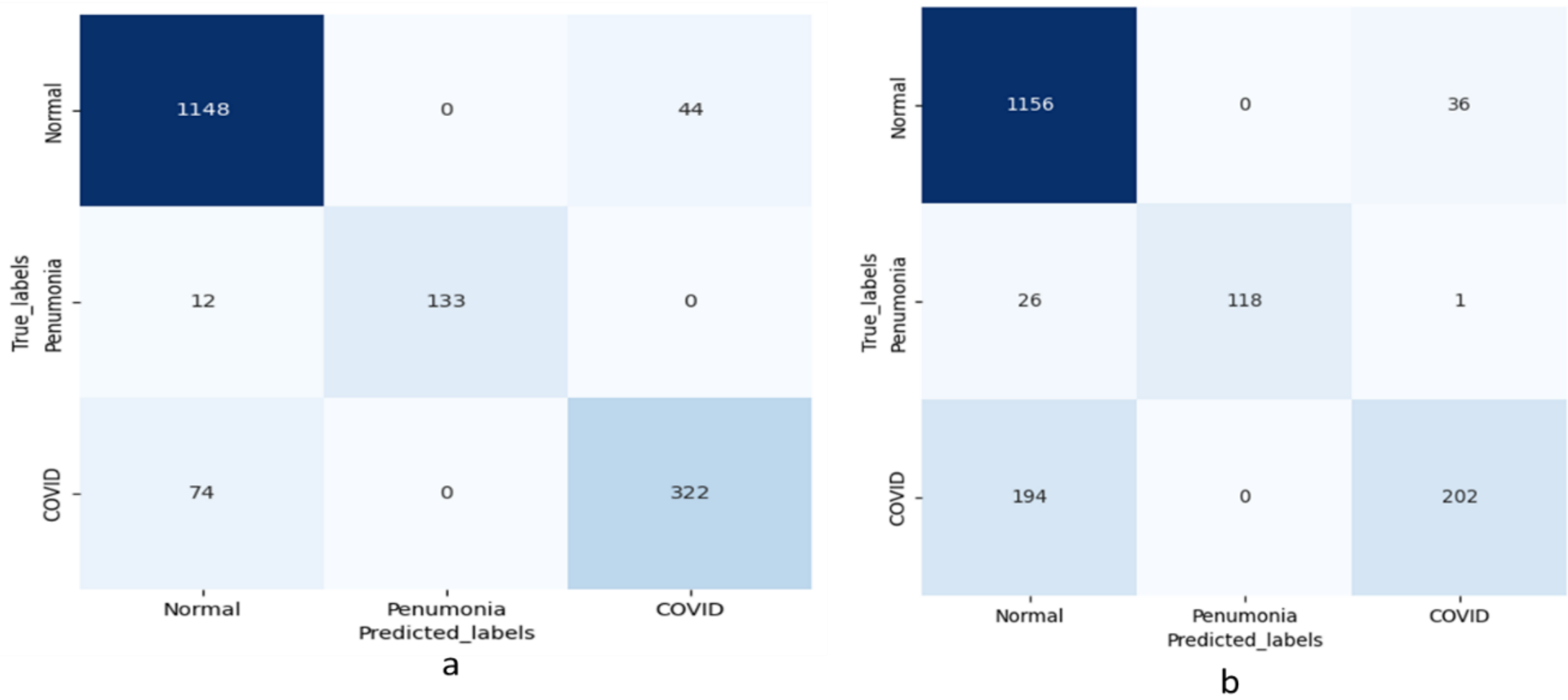

Fig. 9: VGG19 model confusion matrix (a) without augmentation and (b) with augmentation

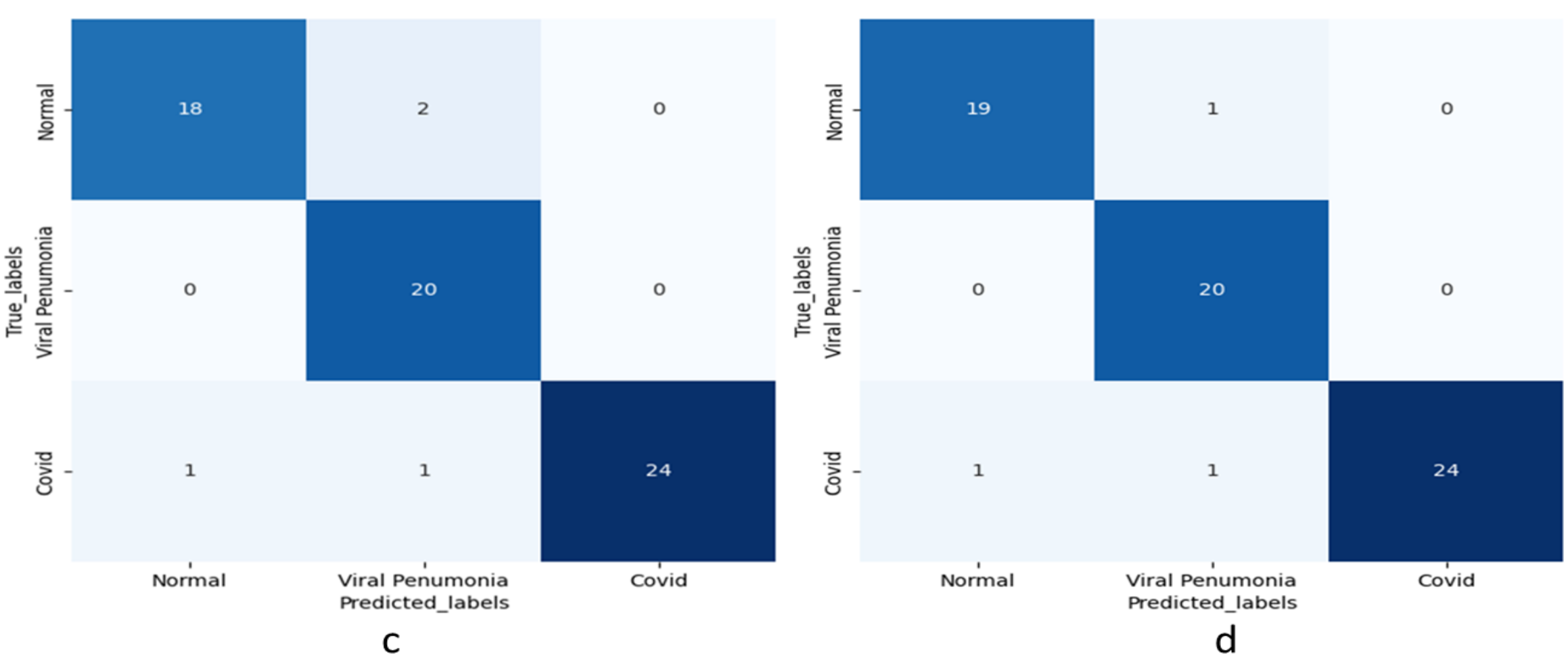

Fig. 10: ResNet50v2 model confusion matrix (c) without augmentation and (d) with augmentation

From Table 2 and Table 3, it can be easily observed that the accuracy of VGG16, VGG19, and ResNet-50v2 models without augmentation are 96%, 92%, and 93%, respectively, while the same for those with augmentation are 94%, 85%, and 95%, respectively.

## 5 CONCLUSION AND FUTURE SCOPE

The fusion of deep learning and medical imaging is an exciting field that can bring revolutionary advancements in disease diagnosis and treatment. The VGG16, VGG19, and ResNet-50v2 models have been used in this paper to classify lung diseases through X-ray images and have shown good results. These deep learning models were characterized by high accuracy, sensitivity, and specificity, and their effectiveness in automating the detection and classification of various pulmonary conditions. With its residual network architecture, ResNet-50v2 demonstrated the possibility of having advantages in the capture of complex features that are important in the diagnosis of an illness. The research is significant because it has resulted in an efficient approach towards the automation of the classification of lung diseases.

In future work, research will focus on optimizing model parameters, using ensemble techniques, improving interpretability, and advancing transfer learning. Generalization is achieved by including many instances and rare instances in the dataset. The practical implementation requirements include addressing the need for solving issues such as integrating the technique into the healthcare system and the compliance issue.

## REFERENCE

1. Niehues, S. M., Adams, L. C., Gaudin, R. A., Erxleben, C., Keller, S., Makowski, M. R., ... &Bressem, K. K. (2021). Deep-learning-based diagnosis of bedside chest X-ray in intensive care and emergency medicine. Investigative Radiology, 56(8), 525-534.
2. Gonem, S., Janssens, W., Das, N., & Topalovic, M. (2020). Applications of artificial intelligence and machine learning in respiratory medicine. Thorax, 75(8), 695-701.
3. Jogin, M., Madhulika, M. S., Divya, G. D., Meghana, R. K., & Apoorva, S. (2018, May). Feature extraction using convolution neural networks (CNN) and deep learning. In 2018 3rd IEEE international conference on recent trends in electronics, information & communication technology (RTEICT) (pp. 2319-2323). IEEE.
4. Suryawanshi, V., Adivarekar, S., Bajaj, K., & Badami, R. (2023, January). Comparative Study of Regularization Techniques for VGG16, VGG19 and ResNet-50 for Plant Disease Detection. In International Conference on Communication and Computational Technologies (pp. 771-781). Singapore: Springer Nature Singapore.
5. Sekine, Y., Katsura, H., Koh, E., Hiroshima, K., & Fujisawa, T. (2012). Early detection of COPD is important for lung cancer surveillance. European Respiratory Journal, 39(5), 1230-1240.
6. Rabe, K. F., Hurd, S., Anzueto, A., Barnes, P. J., Buist, S. A., Calverley, P., ... & Zielinski, J. (2007). Global strategy for the diagnosis, management, and prevention of chronic obstructive pulmonary disease: GOLD executive summary. American journal of respiratory and critical care medicine, 176(6), 532-555.
7. Karar, M. E., Hemdan, E. E. D., & Shouman, M. A. (2021). Cascaded deep learning classifiers for computer-aided diagnosis of COVID-19 and pneumonia diseases in X-ray scans. Complex & Intelligent Systems, 7, 235-247.
8. Halalli, B., &Makandar, A. (2018). Computer aided diagnosis-medical image analysis techniques. Breast imaging, 85, 85-109.
9. Litjens, G., Kooi, T., Bejnordi, B. E., Setio, A. A. A., Ciompi, F., Ghafoorian, M., ... & Sánchez, C. I. (2017). A survey on deep learning in medical image analysis. Medical image analysis, 42, 60-88.
10. LeCun, Y., Bengio, Y., & Hinton, G. (2015). Deep learning. nature, 521(7553), 436-444.
11. Rajpurkar, P., Irvin, J., Ball, R. L., Zhu, K., Yang, B., Mehta, H., ... & Lungren, M. P. (2018). Deep learning for chest radiograph diagnosis: A retrospective comparison of the CheXNeXt algorithm to practicing radiologists. PLoS medicine, 15(11), e1002686.
12. Simonyan, K., & Zisserman, A. (2014). Very deep convolutional networks for large-scale image recognition. arXiv preprint arXiv:1409.1556.
13. He, K., Zhang, X., Ren, S., & Sun, J. (2016). Deep residual learning for image recognition. In Proceedings of the IEEE conference on computer vision and pattern recognition (pp. 770-778).
14. Irvin, J., Rajpurkar, P., Ko, M., Yu, Y., Ciurea-Ilcus, S., Chute, C., ... & Ng, A. Y. (2019, July). Chexpert: A large chest radiograph dataset with uncertainty labels and expert comparison. In Proceedings of the AAAI conference on artificial intelligence (Vol. 33, No. 01, pp. 590-597).
15. Khan, I.U.; Aslam, N. A Deep-Learning-Based Framework for Automated Diagnosis of COVID-19 Using X-ray Images. Information 2020, 11, 419.

16. Esteva, A., Kuprel, B., Novoa, R. A., Ko, J., Swetter, S. M., Blau, H. M., & Thrun, S. (2017). Dermatologist-level classification of skin cancer with deep neural networks. nature, 542(7639), 115-118.
17. He, K., Zhang, X., Ren, S., & Sun, J. (2016). Deep residual learning for image recognition. In Proceedings of the IEEE conference on computer vision and pattern recognition (pp. 770-778).
18. Alshmrani, G. M. M., Ni, Q., Jiang, R., Pervaiz, H., &Elshennawy, N. M. (2023). A deep learning architecture for multi-class lung diseases classification using chest X-ray (CXR) images. Alexandria Engineering Journal, 64, 923-935.
19. Fawzi, A., Samulowitz, H., Turaga, D., & Frossard, P. (2016, September). Adaptive data augmentation for image classification. In 2016 IEEE international conference on image processing (ICIP) (pp. 3688-3692). Ieee.
20. Yang, H., Ni, J., Gao, J., Han, Z., & Luan, T. (2021). A novel method for peanut variety identification and classification by Improved VGG16. Scientific Reports, 11(1), 15756.
21. Dey, N., Zhang, Y. D., Rajinikanth, V., Pugalenthi, R., & Raja, N. S. M. (2021). Customized VGG19 architecture for pneumonia detection in chest X-rays. Pattern Recognition Letters, 143, 67-74.
22. Prusty, S., Patnaik, S., & Dash, S. K. (2022, August). ResNet50V2: A Transfer Learning Model to Predict Pneumonia with chest X-ray images. In 2022 International Conference on Machine Learning, Computer Systems and Security (MLCSS) (pp. 208-213). IEEE.
23. Halder, A., & Datta, B. (2021). COVID-19 detection from lung CT-scan images using transfer learning approach. Machine Learning: Science and Technology, 2(4), 045013.
24. Syed, A. H., Khan, T., & Khan, S. A. (2023). Deep Transfer Learning Techniques-Based Automated Classification and Detection of Pulmonary Fibrosis from Chest CT Images. Processes, 11(2), 443.
25. Koço, S., & Capponi, C. (2013, October). On multi-class classification through the minimization of the confusion matrix norm. In Asian Conference on Machine Learning (pp. 277-292). PMLR.
26. Susmaga, R. (2004). Confusion matrix visualization. In Intelligent Information Processing and Web Mining: Proceedings of the International IIS: IIPWM '04 Conference held in Zakopane, Poland, May 17–20, 2004 (pp. 107-116). Berlin, Heidelberg: Springer Berlin Heidelberg.
27. Hossin, M., & Sulaiman, M. N. (2015). A review on evaluation metrics for data classification evaluations. International journal of data mining & knowledge management process, 5(2), 1.
28. Ozturk, T., Talo, M., Yildirim, E. A., Baloglu, U. B., Yildirim, O., & Acharya, U. R. (2020). Automated detection of COVID-19 cases using deep neural networks with X-ray images. Computers in biology and medicine, 121, 103792.
29. Grandini M, Bagli E, Visani G. Metrics for multi-class classification: an overview. arXiv preprint arXiv:2008.05756. 2020 Aug 13.